# A Holistic Approach for Modeling and Synthesis of Image Processing Applications for Heterogeneous Computing Architectures

Christian Hartmann, Marc Reichenbach, Dietmar Fey
Chair of Computer Architecture
Friedrich-Alexander University Erlangen-Nürnberg (FAU),
Martensstr. 3, 91058 Erlangen, Germany
{christian.hartmann,marc.reichenbach,dietmar.fey}@cs.fau.de

Anna Yupatova, Reinhard German
Chair of Computer Networks and Communication Systems
Friedrich-Alexander University Erlangen-Nürnberg (FAU),
Martensstr. 3, 91058 Erlangen, Germany
{anna.yupatova,reinhard.german}@fau.de

*Abstract*—Image processing applications are common in every field of our daily life. However, most of them are very complex and contain several tasks with different complexities which result in varying requirements for computing architectures. Nevertheless, a general processing scheme in every image processing application has a similar structure, called image processing pipeline: (1) capturing an image, (2) pre-processing using local operators, (3) processing with global operators and (4) post-processing using complex operations. Therefore, application-specialized hardware solutions based on heterogeneous architectures are used for image processing [1]. Unfortunately the development of applications for heterogeneous hardware architectures is challenging due to the distribution of computational tasks among processors and programmable logic units. Nowadays, image processing systems are started from scratch which is time-consuming, error-prone and inflexible. A new methodology for modeling and implementing is needed in order to reduce the development time of heterogenous image processing systems. This paper introduces a new holistic top down approach for image processing systems. Two challenges have to be investigated. First, designers ought to be able to model their complete image processing pipeline on an abstract layer using UML. Second, we want to close the gap between the abstract system and the system architecture.

## I. INTRODUCTION

Setting up an embedded application, which uses high performance image processing on heterogeneous computer architectures is a very complex task. In the traditional industrial image processing field, engineers follow Moore's Law and use standard CPUs for their image processing applications. This solution is not resource aware and does not work for embedded applications. Due to continuous rising requirements on the one hand, and physical limitations of embedded applications concerning area, time and energy, embedded image processing systems become more heterogeneous for fulfilling their functions. These restrictions do not allow the use of oversized general purpose hardware architectures. That leads to the approach of using more application-specialized computing architectures like GPUs or own specialized circuits for FPGAs. The design flow of mapping complex image processing applications on heterogenous architectures is a tough challenge and not sufficient solved in the past. Complex image processing algorithms with different tasks in different granularity have different hardware requirements. These algorithms have to utilize the advantages of specialized hardware architectures for fulfilling the system constraints. Therefore, not only software engineers, but especially hardware engineers, application engineers and system designers are needed, in order to cover all parts of such a system development. Regular programming languages like C/C++ or hardware description languages like VHDL are not applicable for that, because they do not promote a holistic view of the system development. These languages lead to a strict separation of software and hardware and do not consider evaluation techniques for the whole system.

## II. IMAGE PROCESSING DESIGN FLOW

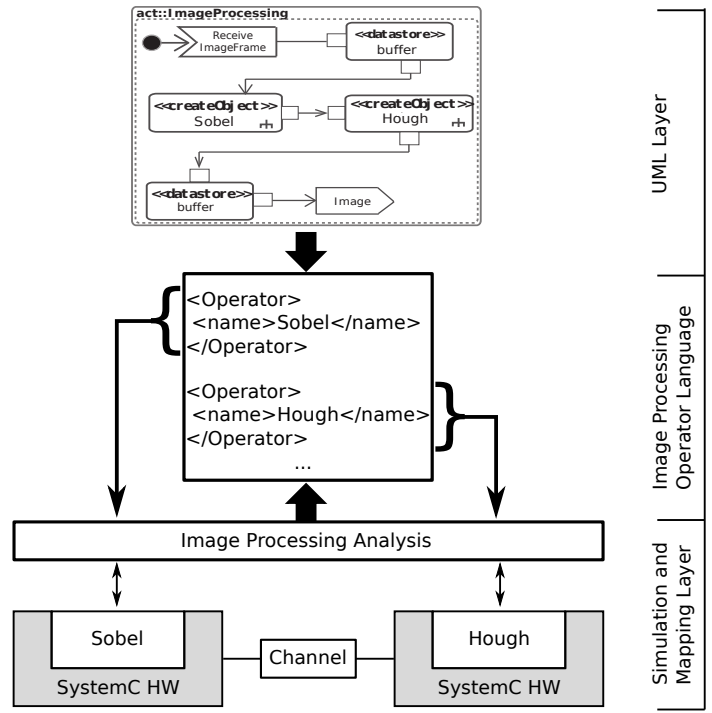

Fig. 1. Image processing design flow: The design flow enables the possiblity to create an image processing application using UML. The UML model will be automatically transfered in an executable SystemC simulation. This is done by the rules of the domain-specific language Image Processing Operator Language (IPOL) and controlled by the Image Processing Analysis.

This work introduces a new system design workflow for image processing systems. The design flow will cover all the aspects of system design, to consider non-functional properties in the abstract design layers. The design process is based on the image processing domain comprising components like image processing algorithms, image sources like image sensors and image sinks like displays which are specified in an intermediate representation level, an own domain-specific language, called Image Processing Operator Language (IPOL). This representation allows an abstract and programming language independent development. However it is bound to the image processing domain. This results, for example, in so-called *properties*: *sensor resolution* and *frames per second*. The new structual features provide the basis for later mapping into concrete hardware and tracing back hardware features to the UML. Figure 1 shows the concept of an image processing design flow. It affords the user the possibility for developing the image processing application in an abstract layer like UML. An automated mapping of the image processing operators to the simulation environment with virtual hardware makes an holistic UML based design approach feasible. Thus the user is able to design an image processing application without detailed knowledge of hardware architecture.

## III. IMAGE PROCESSING OPERATOR LANGUAGE (IPOL)

By using Model-To-Text-Transformation (M2T) [2] the UML model will be transfered in a domain-specific language called Image Processing Operator Language (IPOL). IPOL is a XML based language. It specifies image processing algorithms as image processing operators with hardware influencing properties. With the IPOL language a new domain-specific language for the automated derivation of image processing applications was created. In Listing 1 the structure of IPOL is demonstrated. The example shows the whole image processing application named "operatorchain", with a sensor and one image processing operator. The properties "input_area" and "output_area" of the image processing operator "Sobel" demonstrates its input and output behavior. The "base_calc" block includes a formal description of the algorithm. In this paper "base_calc" remain omitted. In the example of Listing 1 the image processing operator named Sobel needs a $3 \times 3$ neighborhood of picture elements for the calculation of an $1 \times 1$ output region. If we compare the behavior of the Sobel with

```
<operatorchain>
<image_operator id="0">
  <type>Sensor</type>
  <res><x>1920</x><y>1080</y></res>
  <pixres>12</pixres>
  <fps>30</fps>
</image_operator>

<image_operator id="1">
  <type>Operator</type>
  <name>Sobel</name>
  <input_area>
    <x>3</x><y>3</y>
  </input_area><base_calc>...</base_calc>
  <output_area>
    <x>1</x><y>1</y>
  </output_area>
</image_operator>
...
<connections id="0">
  <con id="0"><out>0</out><in>1</in></con>
```

```
</connections>
</operatorchain>
```

Listing 1. Example of an operator chain in the image processing language

an other image processing algorithm, e.g. the Hough Transformation [3], a different input and output behavior will be observed. For unknown image processing algorithms our approach provides a library of common image processing operations, like matrix operations. These library could be used for creating new custom image processing algorithms. The known behavior of the common image processing operations makes it possible to analyse these kind of algorithm without knowing the algorithm itself. The different behavior is used in the SystemC based simulation and analysis environment for calculating the memory bandwidth requirements of the image processing operator shown in Figure 1. If an image processing system is specified in such a language, an automatic derivation to hard- and software components becomes possible. This approach was often discussed in Hardware-Software Co-Design publications [4] and [5] but investigated only small grained architectures with static solutions. Here, we wanted to propose an additional approach for whole image processing systems, with heterogenous architectures consisting of complete processing units like CPU cores, GPUs or special architectures on FPGAs.

## IV. CONCLUSION AND FUTURE WORK

In this paper, a new concept and first steps of the underlying methodology have been introduced for modeling and implementing complex image processing systems with a holistic top down approach on heterogenous computer architectures. The approach covers all layers from abstract UML to a domain-specific language to the executable specification with virtual hardware down to real hardware. In one of our next research steps, we are going to design a connection to the Open Virtual Platform (OVP) [6] and SoClib [7]. That will help developers to include the simulation results for specific processor architectures in their model, without the need of possessing that processor in physical.

## ACKNOWLEDGMENT

This work was financially supported by the Research Training Group 1773 "Heterogeneous Image Systems", funded by the German Research Foundation (DFG).